\documentclass[11pt, copyright, logo, secondlogo]{template/lark}
\usepackage[authoryear, sort&compress, round]{natbib}
\usepackage{bbm}
\usepackage{enumitem}
\setlist{nosep}
\usepackage{xspace}
\usepackage{graphicx}
\usepackage{tabularx}
\usepackage{booktabs}
\usepackage{subcaption}
\usepackage{xcolor}
\usepackage{hyperref}
\usepackage{verbatim}
\usepackage{float}
\usepackage{cleveref}
\usepackage{wrapfig}
\usepackage{makecell}
\usepackage{multirow}
\usepackage{fancyvrb}
\usepackage[normalem]{ulem}
\usepackage{url}
\usepackage{tcolorbox}
\usepackage{listings}
\usepackage{multicol}
\usepackage{diagbox}
\tcbuselibrary{listings,skins,breakable}
\usepackage{algpseudocode}
\usepackage{longtable,array,ragged2e}
\definecolor{darkred}{RGB}{180, 45, 45}
\newtcolorbox{promptbox}[2][Prompt]{
colback=black!5!white,
arc=5pt, 
boxrule=0.5pt,
fonttitle=\bfseries,
title=#1, 
before upper={\small}, fontupper=\fontfamily{ptm}\selectfont,
colframe=#2,
}
\newcommand{\method}{\textsc{Mosaic}}
\newcommand{\monospace}[1]{{\fontfamily{qcr}\selectfont #1}}

\usepackage{amsmath,amsfonts,bm}

\def\eqref#1{equation~\ref{#1}}
\def\1{\bm{1}}

\def\vtheta{{\bm{\theta}}}

\DeclareMathAlphabet{\mathsfit}{\encodingdefault}{\sfdefault}{m}{sl}
\SetMathAlphabet{\mathsfit}{bold}{\encodingdefault}{\sfdefault}{bx}{n}

\title{\method{}: Multi-Session Agentic Reinforcement Learning}

\author[1]{Yilong Xu}
\author[1]{Minrui Xu}
\author[1]{Xinyu Zhou}
\author[2,3$\dag \ddagger$]{Xiang Long, Haochen Tan}
\author[1,4$\ddagger$]{Zhijiang Guo}

\affil[1]{LARK, HKUST(GZ)}
\affil[2]{Foundation Model Lab, Huawei}
\affil[3]{City University of Hong Kong}
\affil[4]{HKUST}

\correspondingauthor{Zhijiang Guo (zhijiangguo@hkust-gz.edu.cn), Haochen Tan (haochetan2-c@my.cityu.edu.hk)}
\githubpage{https://lark-ai-lab.github.io/Mosaic/}

\begin{abstract}

Most existing agentic reinforcement learning methods assume single-session trajectories, where interaction proceeds within one continuously accumulated context. However, agentic techniques such as context compaction and subagent forking may produce trajectories spanning multiple sessions, creating a mismatch between the single-session training assumption and inference-time execution. Therefore, we introduce \textbf{Multi-Session Agentic Reinforcement Learning}, a more general setting that supports an arbitrary number of sessions within a trajectory. Under this setting, a widely adopted method is to broadcast trajectory-level advantages to all sessions, but this does not distinguish contributions from various sessions or guide session allocation within trajectories. To address this issue, we propose \textbf{\method{}}, a multi-session agentic RL method with critic-free hierarchical optimization. At the trajectory level, it incorporates session-usage signals into advantage estimation. At the session level, it shapes each session’s advantage according to the estimated marginal contribution. Experiments on BrowseComp with Qwen3.5-4B show that \method{} outperforms broadcasting in both context compaction and subagent forking scenarios while using fewer sessions on average. Further analyses show that our method achieves improvements in session utility and allocation, which we suggested is a promising target for future advances in multi-session agentic RL.

\end{abstract}

\begin{document}
\maketitle

\section{Introduction}

Agentic Reinforcement Learning \citep[Agentic RL;][]{zhang2025landscape, jin2025search, gao2025beyond} trains Large Language Model (LLM) agents \citep{yao2022react, shinn2023reflexion} for multi-turn reasoning, tool use, and interaction with external environments via online rollouts and reward feedback, yielding broad performance gains. More recently, LLM agents have incorporated advanced inference techniques, such as context management \citep{mei2025survey,zhang2026agentic} and harness engineering \citep{zhou2026externalization,lin2026agentic}, enabling agentic workflows to tackle increasingly complex and long-horizon tasks. However, most agentic RL methods assume a single-session setting, where each rollout unfolds as a coherent multi-turn trajectory within a continuously accumulated context and yields a single training segment \citep{wang2026information, xu2026envfactory}.

We argue that the single-session setting is not sufficiently general for optimizing LLM agents. For example, context compaction compresses the execution history to free up context space while rewriting the prefix that conditions subsequent actions \citep{nguyen2025sfr, wu2025resum, lu2025scaling}, thereby initiating a new session. Similarly, subagent forking delegates global goal to subagents which creates branch sessions \citep{xu2026self, team2026kimi, song2025llm}. Those resulting trajectories no longer satisfy the single-session assumption, creating a mismatch between training and inference that compromises consistency and limits optimization potential.

\begin{figure}[t]
    \centering
    \includegraphics[width=\linewidth]{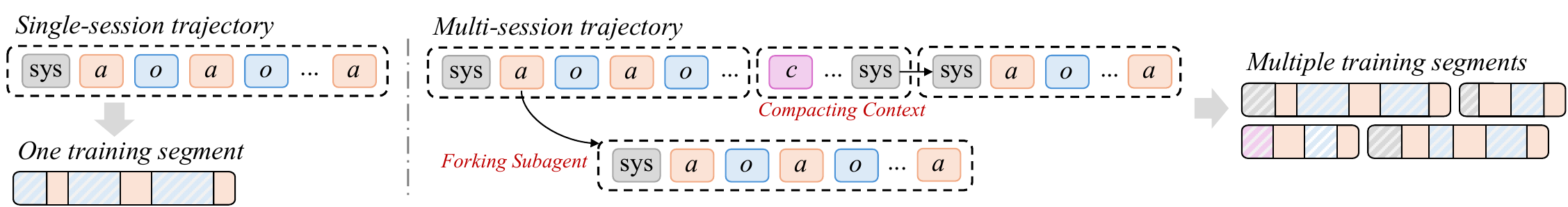}
    \caption{The comparison between multi-session and single-session trajectories.}
    \label{fig:intro}
\end{figure}

To address this gap, we introduce a new setting termed \textbf{Multi-Session Agentic Reinforcement Learning}, which supports an arbitrary count of sessions and enables agents with diverse inference mechanisms to be optimized within a unified algorithm. A straightforward way to handle multi-session rollouts is through \textit{broadcasting}, whereby a trajectory-level outcome reward or advantage is replicated to every token across all sessions \citep{dressage,xu2026polar}. However, we argue that this method provides only coarse supervision and offers no explicit control over optimization across sessions, resulting in poorly targeted learning signals as trajectories span more sessions.

In this work, we propose \textbf{\method{}}, a multi-session agentic RL method with critic-free hierarchical optimization. First, \textbf{session-aware trajectory-level advantage estimation} calibrates outcome-based advantages using within-group rankings of session counts conditioned on task outcomes, encouraging successful trajectories to solve tasks efficiently while allowing unsuccessful trajectories sufficient exploration. Its bounded modulation preserves outcome-induced ordering while incorporating session-usage guidance, thereby widening relative advantage margins for homogeneous-reward groups. Second, \textbf{session-level advantage shaping} calculates each session’s marginal outcome gain as an immediate reward, applies group-relative normalization separately by session type, and combines the resulting shaping signal with the trajectory-level advantage. Overall, the former provides a long-term, outcome-oriented global signal, while the latter captures short-term local marginal effects. They together form a hierarchical optimization target for multi-session agentic RL.

To evaluate our method, we conduct experiments on BrowseComp \citep{wei2025browsecompsimplechallengingbenchmark} and select Qwen3.5-4B \citep{qwen3.5} as the base model. Our \method{} outperforms multiple single-session and multi-session agentic RL baselines under both context compaction and subagent forking scenarios. Moreover, \method{} achieves up to a 10.1\% relative improvement in accuracy over the broadcasting baseline while using fewer sessions per trajectory on average ($3.17 \rightarrow 2.35$). Ablation studies further demonstrate that session-aware trajectory-level advantage estimation and session-level advantage shaping each contribute to the final performance gains. Our analyses also reveal several key insights: (1) shaping trajectory-level advantages with session counts increases within-group advantage variance and improves sample efficiency; (2) different session types exhibit distinct distributions of session-level marginal gains, which our type-conditioned normalization helps stabilize; (3) effective multi-session optimization improves session utility and allocation, a pattern that \method{} strengthens and that might offer a potential direction for future work.

\section{Preliminaries}

\subsection{Agentic RL with Group-Relative Policy Optimization}

Given a task $x\sim\mathcal{D}$, an agent policy $\pi_\vtheta$ generates a rollout trajectory $\tau$ through multi-turn reasoning, tool use, and interaction with the environment. Formally, we represent the trajectory as:
\begin{equation}
    \tau=(a_1,o_1,\ldots,a_T,o_T), \qquad a_t\sim\pi_\vtheta(\cdot\mid h_t),
\end{equation}
where $a_t$ denotes the agent action at turn $t$, $o_t$ is the corresponding observation returned by the environment, and $h_t=\left(x,a_{<t},o_{<t} \right)$ denotes the interaction history available to the policy. Upon completion of the rollout, the environment or a verifier assigns a scalar outcome reward $R \left(\tau \right)$. The objective of Agentic RL is to maximize the expected reward over the task distribution:
\begin{equation}
    \max_\vtheta \mathbb{E}_{x\sim\mathcal{D},,\tau\sim\pi_\vtheta(\cdot\mid x)} \left[R(\tau)\right].
\end{equation}

To estimate policy advantages without learning a separate value function, Group Relative Policy Optimization \citep[GRPO;][]{shao2024deepseekmathpushinglimitsmathematical} uses the relative rewards of multiple rollouts sampled for the same task. Specifically, for each task $x$, GRPO samples a group of rollouts from the old policy $\pi_{\vtheta_{\mathrm{old}}}$: $G(x)=\left\{ \tau_i \right\}_i^G$. Each rollout receives a trajectory-level reward $R_i=R(\tau_i)$. GRPO then estimates its relative advantage using the reward statistics within the group:
\begin{equation}
\hat{A}_i^{\mathrm{traj}}=\frac{R_i - \mu_G}{\sigma_G + \epsilon},
\end{equation}
where $\mu_G$ and $\sigma_G$ are arithmetic mean and standard deviation of rewards in group $G$, respectively. Unlike PPO-style methods \citep{schulman2017proximalpolicyoptimizationalgorithms} that typically rely on a learned critic to estimate token-level or state-level advantages, GRPO constructs its baseline directly from group-level reward statistics. This critic-free design reduces the computational and memory overhead of training an additional value model and avoids errors arising from inaccurate value estimation.

Ignoring optional KL regularization for clarity, the GRPO objective can be written as:
\begin{equation}
    \mathcal{J}_{\text{GRPO}} = \mathbb{E}_{x \sim D, \left\{\tau_i\right\}_{i=1}^G \sim \pi_{\vtheta_{\mathrm{old}}} \left( \tau | q \right)} \frac{1}{G} \sum_{i=1}^{G} \frac{1}{|\tau_i|} \sum_{t=1}^{|\tau_i|} \min \left( \rho_{i,t}(\vtheta)\hat{A}_i^{\mathrm{traj}}, \mathrm{clip}(\rho_{i,t}(\vtheta))\hat{A}_i^{\mathrm{traj}} \right),
\end{equation}
where $\rho_{i,t}=\frac{\pi_\vtheta(a_{i,t}\mid h_{i,t})}{\pi_{\vtheta_{\mathrm{old}}}(a_{i,t}\mid h_{i,t})}$ is the token-level importance ratio.

Most existing GRPO variants, however, assume a single-session setting in which each rollout is represented as a continuous, linear training segment. This assumption becomes inadequate when advanced inference workflows extend agent execution across multiple sessions, creating a mismatch between training and inference that compromises consistency and limits optimization potential.

\subsection{Broadcasting for Multi-Session Agentic RL}

We introduce \textit{session} as a finer-grained unit within an LLM agent trajectory. Specifically, a trajectory $\tau_i$ is represented as a set of $N_i \in \mathbb{N}$ of sessions:
\begin{equation}
    \tau_i = \left \{ s_{i,j} \right \}_{j=1}^{N_i}.
\end{equation}
Each session consists of a sequence of agent–environment interactions same as trajectories in conventional single-session settings:
\begin{equation}
    s_{i,j}=(a_{i,j,1},o_{i,j,1},\ldots,a_{i,j,T_{i,j}},o_{i,j,T_{i,j}}), \qquad a_{i,j,t}\sim\pi_\vtheta(\cdot\mid h_{i,j,t}).
\end{equation}
For example, a session may include consecutive turns of execution to complete a task. When the remaining context window becomes insufficient, a new session can be initiated to perform context compaction, with the resulting compressed prefix serving as the initial context for another session.

Built upon the GRPO pipeline, \textit{broadcasting} makes multi-session rollouts trainable by propagating the trajectory-level advantage $\hat{A}_i^{\mathrm{traj}}$ to every session and every trainable token within $\tau_i$:
\begin{equation}
    \hat{A}_{i,j,t}^{\mathrm{token}} = \hat{A}_{i,j}^{\mathrm{session}}, \enspace \hat{A}_{i,j}^{\mathrm{session}} = \hat{A}_{i}^{\mathrm{traj}}, \quad \forall s_{i,j} \in \tau_i, \enspace t \in \left\{t \mid 1 \le t \le T_{i,j} \wedge \texttt{mask}(t)=1 \right\}.
\end{equation}
Since each session corresponds to a separate training segment in frameworks such as Megatron \citep{shoeybi2020megatronlmtrainingmultibillionparameter}, the token-level objectives from all sessions should further be aggregated at the trajectory level:
\begin{equation}
\label{equation:broadcast_objective}
\begin{aligned}
    {\mathcal{J}}'_{\text{GRPO}} =& \mathbb{E}_{x \sim D, \left\{\tau_i\right\}_{i=1}^G \sim \pi_{\vtheta_{\mathrm{old}}} \left( \tau | q \right)} \frac{1}{G} \sum_{i=1}^{G} {\color{darkred} \frac{1}{\sum_{j=1}^{N_i}|s_{i,j}|} \sum_{j=1}^{N_i} \sum_{t=1}^{|s_{i,j}|} } \min \left( \rho_{i,t}(\vtheta){\color{darkred} \hat{A}_{i,j,t}^{\mathrm{token}}}, \mathrm{clip}(\rho_{i,t}(\vtheta)){\color{darkred} \hat{A}_{i,j,t}^{\mathrm{token}}} \right).
\end{aligned}
\end{equation}

This formulation yields a single token-weighted trajectory-level objective for each rollout, regardless of its session count, thereby preserving equal weighting across rollouts within a prompt group and stabilizing training-signal scales across tokens. However, broadcasting offers only coarse supervision and no explicit control over optimization across sessions, causing poorly targeted learning signals as trajectories span more sessions. To address this limitation, we propose \textbf{\method{}}, which improves multi-session agentic RL through session-aware trajectory-level advantage estimation and session-level advantage shaping.

\begin{figure}[t]
    \centering
    \includegraphics[width=\linewidth]{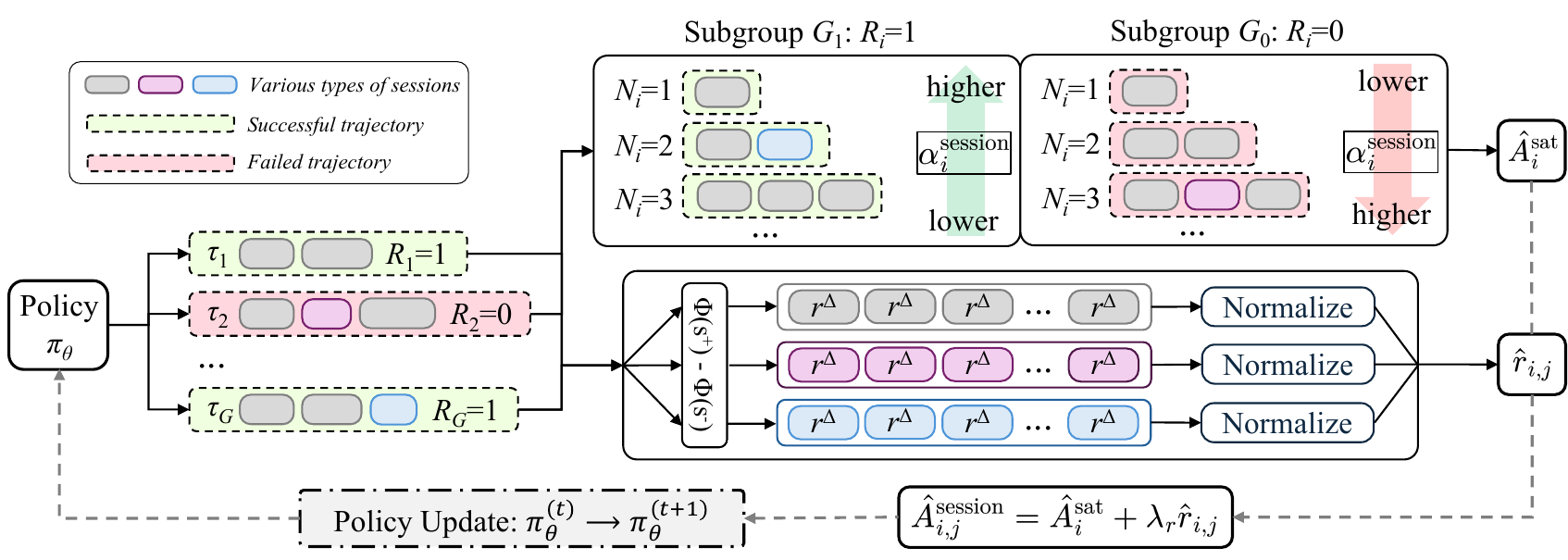}
    \caption{Overview of our proposed \method{} pipeline for multi-session agentic RL.}
    \label{fig:teaser}
\end{figure}

\section{Methodology}

In \method{}, we optimize the performance of multi-session agentic RL through a hierarchical perspective, incorporating trajectory-level advantage estimation calibrated with session signals (\S \ref{method_1_session_aware_traj_adv}), as well as advantage shaping at the finer-grained session level. (\S \ref{method_2_adv_shaping}).

\subsection{Session-Aware Advantage Estimation}
\label{method_1_session_aware_traj_adv}

In vanilla GRPO, the advantage is determined solely by the relative outcome rewards within a prompt group. Under the multi-session setting, however, session-related metrics may provide additional relative signals for policy optimization. One metric is the number of sessions $N_i$ in a trajectory $\tau_i$, which serves as a proxy for the extent of exploration and structural complexity of $\tau_i$. Nonetheless, $N_i$ should not be directly treated as a reward. First, its raw scale may not faithfully quantify meaningful differences in session utilization. Second, the preferred direction of optimization should depend on task results, avoiding the policy from exploiting session count without improving performance.

Accordingly, from a group-relative perspective, the session-aware trajectory-level advantages of multi-session rollouts within a prompt group should satisfy the following ordering constraints:
\begin{itemize}[leftmargin=*, itemsep=0pt, topsep=2pt, parsep=0pt]
    \item \textbf{C1}: For successful trajectories, using more sessions incurs a higher execution cost and hence results in a lower advantage, whereas using fewer sessions should result in a higher advantage.
    \item \textbf{C2}: For unsuccessful trajectories, using fewer sessions may suggest insufficient exploration or execution, especially on challenging tasks, and should thus result in a lower advantage. Conversely, unsuccessful trajectories that conduct more sessions should receive a higher advantage.
    \item \textbf{C3}: Task outcome reward must remain the dominant signal. When two trajectories have the same session counts, the trajectory with lower reward should receive a lower advantage. Using more sessions alone should not allow an unsuccessful trajectory to outrank a successful one.
\end{itemize}

Together, these constraints encourage the policy to use enough sessions to complete the task, but no more than necessary, which represents a desirable objective for multi-session agentic RL. Specifically, we assume a binary final outcome reward $R_i \in \left\{0,1\right\}$, where $1$ and $0$ represents success and failure, respectively. For a prompt group containing both successful and unsuccessful rollouts, the trajectory-level GRPO advantage satisfies:
\begin{equation}
    R_{i}=1 \Longrightarrow \hat{A}_{i}^{\mathrm{traj}}>0,\quad R_{i}=0 \Longrightarrow \hat{A}_{i}^{\mathrm{traj}}<0.
\end{equation}

Based on the ordering constraints above, we define the session-aware trajectory-level advantage as:
\begin{equation}
\label{equation:traj_adv}
    \hat{A}_{i}^{\mathrm{sat}}=\left ( 1+ \alpha^{\mathrm{session}}_i \right ) \hat{A}_{i}^{\mathrm{traj}},
\end{equation}
where $\alpha_i^{\mathrm{session}}$ is the session signal derived from $N_i$, acting as a factor that modulates the trajectory-level advantage according to session usage, and extending optimization beyond the final outcome.

Considering that session signal has different implications under different task outcomes as reflected by C1 and C2, session signals are not directly comparable between successful and unsuccessful trajectories. We thus further divide the prompt group $G$ into subgroups according to $R_i$:
\begin{equation}
G_1=\left \{ i \in G : R_i=1 \right \}, \quad G_0=\left \{ i \in G : R_i=0 \right \}
\end{equation}
Generally, each $\tau_i$ belongs to $G_{R_i}$. We then map its $N_i$ to a centered rank within this subgroup:
\begin{equation}
    u_i=2f(N_i)-1, f(N_i)= \frac{\mathrm{rank}_{G_{R_i}}(N_i)-1}{|G_{R_i}|-1},
\end{equation}
where $u_i \in \left[ -1,1 \right]$, and the rank is computed in ascending order of $N_i$ inside $G_{R_i}$. Thus, a larger $N_i$ yields a larger $u_i$. We use centered ranks instead of raw session counts to obtain a bounded signal robust to long-tailed trajectories, as differences in $N_i$ may not reflect proportional differences in session usage. We then define the session modulation factor as:
\begin{equation}
    \alpha^{\mathrm{session}}_i=- \rho u_i,
\end{equation}
where $0\le\rho<1$ and $\rho$ controls the proportion by which the session signal may amplify or attenuate the original advantage. Then we have $ 1 - \rho u_i \in \left [ 1-\rho, 1+\rho \right ]$ which is always positive. The session signal thus modulates only the magnitude of the outcome advantage without changing its sign.

Finally, we consider the fallback cases in the above computation. First, the centered-rank signal $u_i$ is well-defined only when $\left | G_{R_i} \right | > 1 \wedge \mathrm{Var}_{G_{R_i}} > 0$. If either condition is not satisfied, we set $u_i=0$. When all trajectories in a prompt group receive the same outcome reward, the outcome advantage in Eq. \ref{equation:traj_adv} becomes 0, rendering the session signal ineffective. To improve sample efficiency in these homogeneous groups, we introduce the following fallback in the final advantage:
\begin{equation}
    \boxed{\hat{A}_{i}^{\text {sat}}=\left\{\begin{array}{ll}
\left(1-\rho u_{i}\right) \hat{A}_{i}^{\text {traj }}, & 0<\sum_{j \in G} R_{j}<|G|, \\
-\beta_{\text {1 }} u_{i}, & \sum_{j \in G} R_{j}=|G|, \\
\beta_{\text {0 }} \min \left(u_{i}, 0\right), & \sum_{j \in G} R_{j}=0 .
\end{array}\right.}
\label{equation:final_traj_adv}
\end{equation}
For all-failure groups, we adopt the more conservative fallback, ensuring that no unsuccessful trajectory receives a positive advantage. The nonnegative coefficients $\beta_0$ and $\beta_1$ control the strengths of the fallback signals for all-failure and all-success groups, respectively.

\subsection{Session-Level Advantage Shaping}
\label{method_2_adv_shaping}

The session-aware trajectory-level advantage developed in \S \ref{method_1_session_aware_traj_adv} provides an outcome-oriented global optimization signal for the entire multi-session trajectory. At the session level, however, this signal remains sparse since all sessions on the same trajectory share a single scalar advantage. To this end, we introduce session-level advantage shaping. Specifically, we view each session as a state transition as $s_{i,j}=\left( v_{i,j-1} \rightarrow v_{i,j} \right)$, where session $s_{i,j}$ transitions the agent from state $v_{i,j-1}$ to state $v_{i,j}$. Here, $v_{i,j-1}$ and $v_{i,j}$ denote the agent states at the start and end of the $j$-th session, respectively, with particular emphasis on the prefix context visible to the rollout policy. Based on session transition functions, we introduce three representative session types:
\begin{itemize} [leftmargin=*, itemsep=0pt, topsep=2pt, parsep=0pt]
    \item \textbf{Normal session}. The agent performs regular reasoning, tool use, and environment interaction within the current context. The input state is the context available at the start of the session, and its output state is the resulting context after the session is completed.
    \item \textbf{Compaction session}. When the accumulated context approaches the model’s limit, the agent starts a compaction session to compress the context into a shorter prefix. Its input state is the full context before compaction, and its output state is the compressed prefix passed to the next session.
    \item \textbf{Subagent session}. The main agent delegates subtasks to subagents which start new subagent sessions. Its input state is the main-agent context at delegation, and its output state is the updated main-agent context after the returned result has been incorporated.
\end{itemize}

To quantify the benefit of each session transition, we construct a critic-free process supervision signal based on marginal information gain. Specifically, let $y^*_i = \left ( y^*_{i,1}, \ldots, y^*_{i,L_i} \right )$ denote the ground-truth answer to query $x_i$ of trajectory $\tau_i$. We use the rollout policy before the update of each step, denoted by $\pi_{\bar\vtheta}$, and define the answer-likelihood potential of any state $v$ as:
\begin{equation}
    \Phi_{\bar{\vtheta}} \left ( y^*_i \mid x_i, v \right )=\frac{1}{L_i} \sum^{L_i}_{\ell=1} \log \pi_{\bar{\vtheta}} \left ( y^*_{i,\ell} \mid x_i,v,y^*_{i,<\ell} \right ).
\end{equation}
Intuitively, $\Phi_{\bar\vtheta}(y_i^*\mid x_i,v)$ measures how readily the rollout policy can generate the correct answer given the prefix context represented by state $v$. A session that adds useful evidence, reduces noise, or returns a high-quality subtask result should increase the average log-likelihood of the ground-truth answer. Conversely, a transition that discards key facts or introduces misleading information may decrease it. Accordingly, for a session transition $s_{i,j}=\left( v_{i,j-1} \rightarrow v_{i,j} \right)$, we define its session-level reward as the marginal information gain between its output and input states:
\begin{equation}
    r^{\Delta}_{i,j} = \mathrm{stopgrad} \left [ \Phi_{\bar{\vtheta}} \left ( y^*_i \mid x_i, v_{i,j} \right )- \Phi_{\bar{\vtheta}} \left ( y^*_i \mid x_i, v_{i,j-1} \right ) \right ].
\end{equation}
Different types of session may have different distributions of marginal gain. Normal sessions typically gain external evidence through search and browse. The gain from a compaction session depends on whether compression retains useful information while removing noise. A subagent session is affected by both the quality of subtask execution and the quality of result handoff. Therefore, we normalize the session-level rewards separately by session type within each prompt group:
\begin{equation}
\hat{r}_{i,j} = \begin{cases} \frac{r^{\Delta}_{i,j}-\mu_{c\left(s_{i,j}\right)}}{\sigma_{c\left(s_{i,j}\right)}+\epsilon} & \text{ if } 1 \le j < N_i, \\ 0 & \text{ if } j = N_i, \end{cases}
\end{equation}
where $c\left(s_{i,j}\right) \in \left\{ \mathrm{normal}, \mathrm{compact}, \mathrm{subagent}, \dots \right\}$ denotes the type of session $s_{i,j}$, and $\mu_{c(s_{i,j})}$ and $\sigma_{c(s_{i,j})}$ are computed over sessions of the same type within the prompt group. We exclude the last session since its output state usually includes the agent's final answer. Particularly, a correct final answer may sharply increase the answer likelihood which confounds the marginal gain distribution. Setting its shaping signal to 0 prevents the final answer from entering the evaluated output state.

Finally, we combine the normalized session-level shaping signal with the trajectory-level advantage:
\begin{equation}
    \hat{A}_{i,j}^{\mathrm{session}}=\hat{A}^{\mathrm{traj}}_i + \lambda_r \hat{r}_{i,j},
\end{equation}
where $\lambda_r$ controls the intensity of the shaping signal, and $\hat{A}^{\mathrm{traj}}_i$ can be instantiated directly as $\hat{A}^{\mathrm{sat}}_i$ in \S \ref{method_1_session_aware_traj_adv}, after which $\hat A_{i,j}^{\mathrm{session}}$ replaces the broadcast session advantage used to compute $\hat A_{i,j,t}^{\mathrm{token}}$ in Eq. \ref{equation:broadcast_objective}. Overall, the first term provides a long-term, outcome-oriented global signal, while the second differentiates sessions according to their local relative marginal gains.

\begin{table}[t]
\caption{Results of various methods and inference modes on BrowseComp.}
\vspace{5pt}
\label{tb:browsecomp_main}
\resizebox{\linewidth}{!}{
\renewcommand\arraystretch{1.1}

\begin{tabular}{lcccccc}
\toprule
\rowcolor{gray!20} & \multicolumn{2}{c}{\textbf{Single-Session}} & \multicolumn{2}{c}{\textbf{Multi-Session / Compact}} & \multicolumn{2}{c}{\textbf{Multi-Session / Subagent}} \\
\rowcolor{gray!20} \multirow{-2}{*}{\diagbox{\textbf{Method}}{\textbf{Inference}}} & Accuracy & Num. Session & Accuracy & Num. Session & Accuracy & Num. Session \\
\midrule
Qwen3.5-4B & 23.22 & 1.00 & 24.80 & 1.70 & 21.64 & 2.98 \\
\multicolumn{7}{c}{\textcolor{gray}{\textit{Training on Single-Session}}} \\
GRPO & 31.99 & 1.00 & 34.68 & 1.24 & 31.91 & 2.09 \\
IGPO & 28.52 & 1.00 & 31.36 & 4.12 & 24.64 & 3.26 \\
\multicolumn{7}{c}{\textcolor{gray}{\textit{Training on Multi-Session}}} \\
Broadcasting & -- & -- & 35.31 & 3.17 & 31.44 & 4.43 \\
\method{} & -- & -- & \textbf{38.86} & 2.35 & \textbf{32.46} & 3.77 \\
\bottomrule
\end{tabular}
}
\end{table}

\section{Experiments}

\subsection{Experimental Setup}
\label{experimental_setup}

\paragraph{Benchmark.} We evaluate our method on BrowseComp \citep{wei2025browsecompsimplechallengingbenchmark}, a challenging benchmark consisting of 1,266 questions that require agents to persistently navigate the web and identify hard-to-find, entangled information. We select BrowseComp since its high difficulty demands long-horizon interactions, making it suitable for assessing agents that produce multi-session trajectories.

\paragraph{Baselines.} We use Qwen3.5-4B \citep{qwen3.5} as the initial model and adopt a ReAct-style agent loop. We organize the baselines by inference configuration. Under \textbf{single-session inference}, we evaluate (1) the initial model without RL training and (2) a model trained with GRPO in the single-session setting, representing conventional Agentic RL. Under \textbf{multi-session inference}, we evaluate (3) the single-session GRPO trained model, representing test-time scaling; (4) a model trained with Broadcasting in the multi-session setting; and (5) a model trained with \method{} in the multi-session setting. We evaluate all multi-session configurations under both context compaction and subagent forking using their respective agent loops; for multi-session training, the same loops are used to collect rollouts. Finally, since our session-level advantage shaping in \S \ref{method_2_adv_shaping} builds on information gain, we evaluate (6) a single-session model trained with IGPO \citep{wang2026information}.

\paragraph{Implementation Details.} The agent has two tools, \texttt{search} and \texttt{visit}, implemented using Serper \footnote{\url{https://serper.dev/}} and Jina Reader \footnote{\url{https://jina.ai/reader/}} APIs, respectively. During training and evaluation, \texttt{deepseek-v4-flash} \citep{deepseekai2026deepseekv4highlyefficientmilliontoken} serves as the LLM judge for verifying final answers. For context compaction, the agent compresses its interaction history when approaching the context limit. For subagent forking, it can invoke non-nested subagents through a dedicated action. Each session is limited to 128K tokens. For RL training, we use REDSearcher\_RL\_1k \citep{chu2026redsearcherscalablecostefficientframework}, which contains 1,000 challenging deep-search tasks. We train all models for one epoch with a rollout batch size of 16 and a group size of 8. We set $\rho=0.2$, $\beta_0=0.1$, $\beta_1=0.2$. and $\lambda_r=0.1$.

\subsection{RQ1: Overall Effectiveness of Mosaic}

RQ1 asks \textit{whether training in the multi-session setting improves agent performance and whether \method{} contributes beyond broadcasting}.

\begin{figure}[t]
    \centering
    \includegraphics[width=\linewidth]{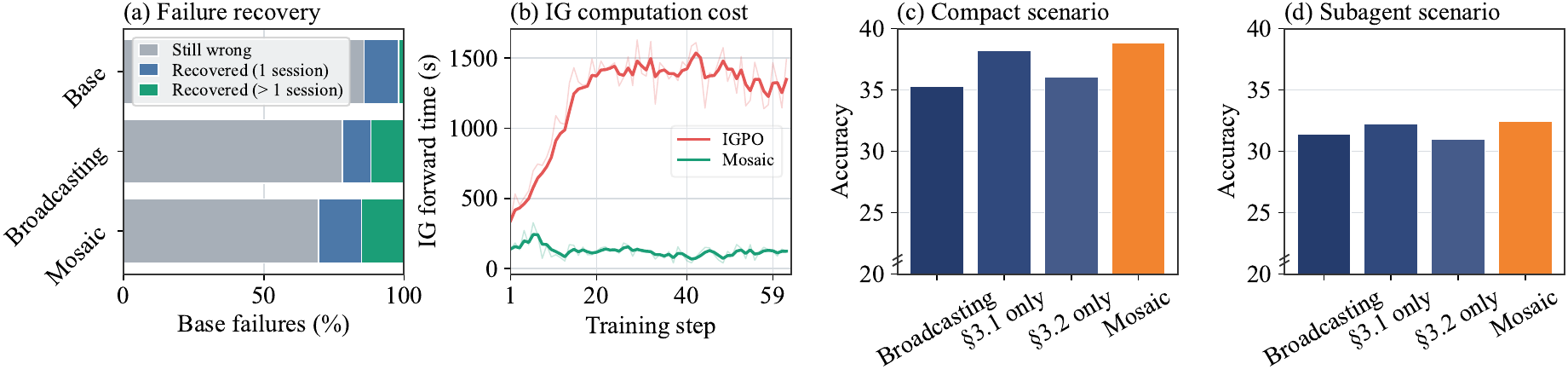}
    \caption{\textbf{(a)}. Percentage of the 972 tasks failed by the base model under single-session inference that are solved by different methods under multi-session inference with compact. \textbf{(b)}. Comparison of the computation time for marginal information gain. \textbf{(c)(d)}. Ablation results on BrowseComp.}
    \label{fig:result_and_ablation}
\end{figure}

As shown in Table \ref{tb:browsecomp_main}, single-session inference has limited effectiveness on BrowseComp. Agentic RL training in the single-session setting yields modest gains, even when combined with test-time scaling in the multi-session setting. This suggests that multi-session inference capabilities do not readily emerge through generalization alone. Among methods trained on multi-session trajectories, Broadcasting improves multi-session inference performance. \method{} improves it further, outperforming all baselines in both the compact and subagent scenarios. Figure \ref{fig:result_and_ablation} (a) illustrates the transformative gains from multi-session inference, multi-session training, and \method{}.

\begin{figure}[t]
    \centering
    \includegraphics[width=\linewidth]{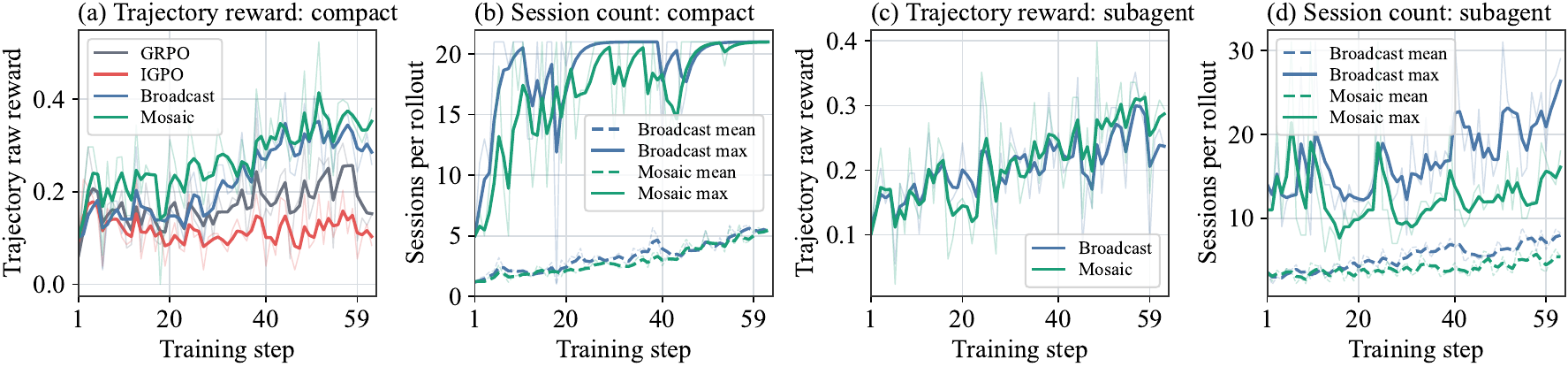}
    \caption{Curves showing rewards and session counts during the training process of various methods.}
    \label{fig:training_dynamic}
\end{figure}

Figure \ref{fig:training_dynamic} shows the training dynamics of each method. \method{} boosts trajectory reward faster than Broadcasting and reaches a higher level. The number of sessions increases over training for both methods, but its growth levels off more clearly with \method{}. This may be partly due to the more targeted signal provided by its session-aware advantage, which discourages unnecessary increases in session count. Overall, \method{} achieves better task performance at lower execution cost.

As \S \ref{method_2_adv_shaping} applies marginal gain to session-level advantage shaping, we thus compare \method{} with IGPO, which applies it at the turn level. From Table \ref{tb:browsecomp_main}, \method{} achieves better performance than IGPO. Moreover, Figure \ref{fig:result_and_ablation} (b) shows that \method{} substantially reduces the wall-time of computing marginal gains, enhancing performance while cutting computational cost.

\subsection{RQ2: Ablation Study of Mosaic}

RQ2 asks \textit{whether each component of \method{} contributes to its overall performance}.

Figure \ref{fig:result_and_ablation} (c)(d) presents the ablation results. Relative to Broadcasting, both levels of components in \method{}, namely trajectory-level session-aware advantage estimation and session-level advantage shaping, contribute to the enhanced final performance. Combining the two components yields a clear additional gain in the compact scenario, but only a modest gain in the subagent scenario. One possible explanation lies in their session structures: compact trajectories are linear, whereas subagent trajectories have a nonlinear structure that more closely resembles a directed acyclic graph. In the latter case, a session may influence multiple succeeding sessions. Its immediate reward, estimated from marginal gain, may thus differ from its actual contribution to the final outcome.

\subsection{RQ3: Mechanistic Analysis of Mosaic}

RQ3 asks \textit{how the design of \method{} contributes to the training process}.

\begin{figure}[t]
    \centering
    \includegraphics[width=\linewidth]{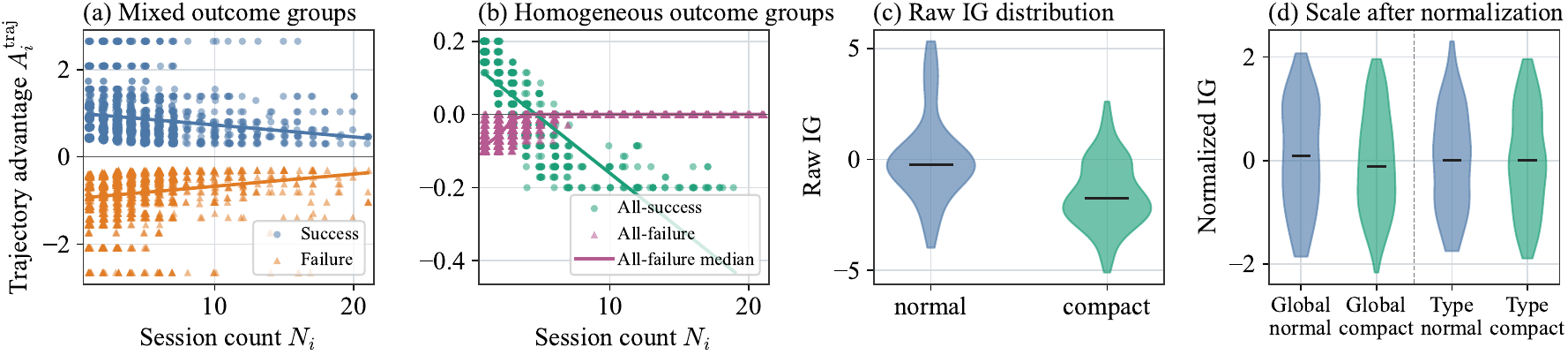}
    \caption{The two left panels show the relationship between trajectory-level advantage and session count. The two right panels show the distributions of marginal information gain for normal and compact sessions under global normalization and type-conditioned normalization.}
    \label{fig:ordering_and_ig_normalization}
\end{figure}

First, we examine whether session-aware trajectory-level advantage estimation produces the intended ordering. As shown in Figure \ref{fig:ordering_and_ig_normalization}, within mixed-outcome groups, advantages approximately decrease with session count for successful trajectories and increase for unsuccessful ones, while successful trajectories retain higher advantages. In homogeneous-outcome groups, the fallback mechanism introduces session-dependent variation, providing differentiated learning signals despite identical outcome rewards. These patterns illustrate how \method{} leverages session signals to increase the within-group variance of relative advantages, thus improving sample efficiency during training.

Second, we examine the effect of reward normalization on session-level shaping signals. As shown in Figure \ref{fig:ordering_and_ig_normalization}, the distributions of raw marginal information gain differ across session types, primarily because heterogeneous sessions serve distinct functions. Under global normalization, compaction sessions, whose raw marginal gains have a lower mean, are further downweighted, introducing a systematic bias against them. In \method{}, type-conditioned normalization instead centers and rescales each type separately, enabling both types to receive positive and negative signals relative to their respective baselines. This reduces type-dependent distributional bias in advantage shaping.

\subsection{RQ4: Understanding Multi-Session Optimization}
\label{rq4}

Multi-session optimization introduces an additional degree of freedom in how computation is distributed across sessions. However, the session count used by \method{} serves only as a proxy signal and is not inherently tied to exploration intensity or execution cost. Moreover, at inference time, the model has no access to hindsight: it does not know how many sessions a trajectory will ultimately contain or whether the trajectory will succeed. Therefore, RQ4 moves beyond method-specific analysis and asks a broader question: \textit{when multi-session optimization improves task performance, what session-level behavioral patterns emerge?}

Accordingly, we examine session-level behavior along three dimensions: \textit{quantity}, or whether the count of sessions increases or decreases; \textit{quality}, or whether performance improvement is accompanied by more useful sessions; and \textit{allocation}, or whether a limited session budget becomes increasingly concentrated on execution stages with high marginal utility. As shown in Table \ref{tb:session_pattern}, for Broadcasting, which does not directly optimize any specific session-aware pattern, all three metrics increase. Under \method{}, the number of sessions does not increase further, while both session quality and allocation improve. These findings indicate that performance gains via multi-session optimization are associated with improved session utility rather than a monotonic rise in session count. Moreover, despite using explicit session-aware optimization, \method{} does not induce a fundamentally altered behavioral mechanism. Instead, it strengthens the session-allocation patterns that already emerge from naïve multi-session optimization.

We argue that effective multi-session optimization is accompanied by improvements in session utility and allocation. It distributes computation more effectively across sessions, encouraging continued execution when it yields meaningful progress while suppressing low-utility computation. From this perspective, the central challenge is not how many sessions an agent should use, but how much useful progress it can derive from the sessions it executes.

\begin{table}[t]
\centering
\caption{Session-level patterns under multi-session optimization on BrowseComp.}
\vspace{4pt}
\label{tb:session_pattern}

\resizebox{\linewidth}{!}{
\renewcommand{\arraystretch}{1.1}
\setlength{\tabcolsep}{6pt}

\begin{tabular}{lcccccccc}
\toprule
& \multicolumn{4}{c}{\textbf{Multi-Session / Compact}}
& \multicolumn{4}{c}{\textbf{Multi-Session / Subagent}} \\
\cmidrule(lr){2-5}
\cmidrule(lr){6-9}

\textbf{Method}
& \textbf{Accuracy} $\uparrow$
& \textbf{\# Sessions}
& \textbf{Positive-IG} $\uparrow$
& \textbf{IG / Session} $\uparrow$
& \textbf{Accuracy} $\uparrow$
& \textbf{\# Sessions}
& \textbf{Positive-IG} $\uparrow$
& \textbf{IG / Session} $\uparrow$ \\
\midrule

Base Model
& 24.80 & 1.70 & 37.36\% & 0.30
& 21.64 & 2.98 & 27.39\% & 0.41 \\

Broadcasting
& 35.31 & 3.17 & 47.74\% & 0.86
& 31.44 & 4.43 & 41.15\% & 0.74 \\

\textbf{\method{}}
& 38.86 & 2.35 & 59.92\% & 0.89
& 32.46 & 3.77 & 51.04\% & 0.80 \\

\bottomrule
\end{tabular}
}
\end{table}

\section{Discussion: Future Directions for Multi-Session Agentic RL}

\paragraph{Infrastructure for reproducible research.} Multi-session trajectories tend to arise naturally when agents tackle complex, long-horizon tasks. Although some existing benchmarks can elicit these trajectories with reasonable frequency, suitable public training data remain scarce. This gap limits the scale, sufficiency, and reproducibility of experiments in multi-session agentic RL. Developing open training datasets and corresponding agentic execution environments is therefore essential to building a sustainable research ecosystem for this topic.

\paragraph{Cross-session state transfer as a central learning objective.} Context compaction and subagent forking both involve information transfer, but serve different purposes: compaction must preserve sufficient working state for continued execution; forking must provide the information needed to complete a subtask; merging must integrate subtask results into a state usable by the main agent. These operations motivate a shared research question: under a limited context budget, what information should be carried forward to support subsequent decisions? CompactionRL already jointly optimizes task execution and context summarization \citep{li2026compactionrlreinforcementlearningcontext}. A unified approach to learning state transfer across compaction, forking, and merging remains an important direction, alongside systematic investigation of information loss and error accumulation over repeated transfers.

\paragraph{Asynchronous training for multi-session rollouts.} For partial rollout or asynchronous training settings, multi-session execution can amplify variation in trajectory lengths and completion times, increase policy staleness and complicate within-group comparisons and sample reuse. For instance, prioritizing completed rollouts may favor trajectories with less sessions over longer ones associated with harder tasks, introducing selection bias and shaping learned execution strategies. Our experiments use a colocated setup with alternating rollout and training phases to minimize interference from asynchrony and isolate the effects of the optimization signals. Scaling toward actual production raises a further question: how can multi-session optimization remain reliable under asynchronous sampling and policy staleness? SAO explores asynchronous optimization with one rollout per prompt, using a learned value model \citep{hou2026singlerolloutasynchronousoptimizationagentic}. However, critic-free, group-relative optimization requires further investigation into when and how rollout groups are formed, how stale samples are handled, and how collected trajectories can be reused.

\paragraph{Toward RL within (black-boxed) agent harnesses.} General multi-session agentic RL can accommodate workflows in which execution no longer follows a single, continuously accumulated context prefix. Context compaction and subagent forking, the two scenarios studied in this work, are representative examples of such workflows in modern agent harnesses. Progress in multi-session optimization could thus provide an important foundation for training agents within harnesses whose orchestration is partially opaque \citep{xu2026polar}, such as Claude Code. Therefore, a critical next step is to systematically examine mechanisms across agent harnesses that disrupt context continuity, such as dropping reasoning tokens or tool responses, and improve general multi-session agentic RL under these conditions. Additionally, when discontinuity occurs, the choice between introducing a new session boundary and applying lossy concatenation with the preceding session may also influence the overall effectiveness of optimization.

\section{Related Work}

\paragraph{LLM Agents.} LLM agents extend LLMs with tool use and environment interaction to work autonomously \citep{feng2026retool,li2025websailor,wu2026webdancer,wang2024videoagent}. ReAct constructs a general agent loop of reasoning, acting, observing environment feedback, and continuing until task completion \citep{yao2022react}. To handle complex and long-horizon tasks, recent systems develop agent harnesses that manage execution beyond the basic ReAct loop \citep{pi-agent,openclaw}, incorporating context compaction \citep{nguyen2025sfr}, multi-agent system \citep{chen2026trainee}, subagent forking \citep{team2026kimi}, and other forms of context and workflow management \citep{shi2026look,xu2026mem,zhang2026agentic,chen2025iterresearch,kim2025towards,yang2026understanding}.

\paragraph{Agentic RL.} Reinforcement learning techniques such as RLHF \citep{ouyang2022training, zhang2026instruction} and RLVR \citep{chen2026does, yu2026dapo} are used to improve LLMs' alignment, reasoning, and downstream task-solving capabilities \citep{zheng2025deepresearcher, tan2025deepscaler}. Agentic RL extends this paradigm to LLM agents, aiming to enhance capabilities of multi-turn interaction through enhanced rollout sampling \citep{dong2026agentic, hou2025treerl}, process-level reward design \citep{zhang2026process, liu2026agentic}, and other optimization strategies \citep{deng2026uncertainty}. However, most existing agentic RL methods still assume that each trajectory is contained within a single session.

\section{Conclusion}

In this paper, we first identify a mismatch between the single-session assumption underlying most existing agentic RL methods and advanced agentic inference techniques such as context compaction and subagent forking. This motivates our formulation of Multi-Session Agentic Reinforcement Learning, which supports an arbitrary count of sessions within a trajectory. To address the coarse supervision issue existed in the currently widely used broadcasting approach, we further propose \method{}, a multi-session agentic RL method with critic-free hierarchical optimization. By combining session-aware trajectory-level advantage estimation with session-level advantage shaping, \method{} trains a Qwen3.5-4B agent that outperforms both single-session and multi-session baselines on BrowseComp. Ablation studies confirm the contribution of each component, while further analyses suggest that effective multi-session optimization might depend on improving session utility and allocation, a pattern observed in both broadcasting baseline and our \method{}. We hope these findings will inform future research on multi-session agentic RL.

\clearpage 

\bibliography{references}
\bibliographystyle{abbrvnat}

\clearpage

\appendix

\section{Details of Section 3.1}

\subsection{Ordering of the Session-Aware Advantage}

We show how the advantage defined in \S \ref{method_1_session_aware_traj_adv} satisfies the ordering constraints C1--C3. Let $G$ be a prompt group, $R_i\in\{0,1\}$ the outcome reward of trajectory $i$, and $N_i$ its number of sessions. Within each outcome subgroup, the centered-rank signal $u_i\in[-1,1]$ increases with $N_i$ whenever the subgroup contains distinct session counts. If a subgroup contains only one trajectory or has zero variance in session count, we set $u_i=0$.

We first consider a mixed-outcome group, for which
$0<\sum_{i\in G}R_i<|G|$. In this case, the session-aware advantage is
\begin{equation}
    \hat A_i^{\mathrm{sat}}
    =(1-\rho u_i)\hat A_i^{\mathrm{traj}},
    \qquad 0<\rho<1,
    \label{eq:app_order_mixed}
\end{equation}
where $\hat A_i^{\mathrm{traj}}$ is the group-relative advantage computed from the binary outcome rewards. Because all trajectories with the same reward have the same outcome advantage, there exist constants $a_+>0$ and $a_-<0$ such that
\begin{equation}
    \hat A_i^{\mathrm{traj}}=
    \begin{cases}
        a_+, & R_i=1,\\
        a_-, & R_i=0.
    \end{cases}
    \label{eq:app_binary_outcome_advantage}
\end{equation}

\paragraph{C1: ordering among successful trajectories.}
Consider two successful trajectories $i$ and $j$ in the same prompt group. If $N_i<N_j$, their within-outcome ranks satisfy $u_i<u_j$. Since both trajectories have outcome advantage $a_+>0$,
\begin{align}
    \hat A_i^{\mathrm{sat}}-\hat A_j^{\mathrm{sat}}
    &=a_+\bigl[(1-\rho u_i)-(1-\rho u_j)\bigr] \notag\\
    &=\rho a_+(u_j-u_i)>0.
    \label{eq:app_c1_proof}
\end{align}
Thus, among successful trajectories, the one using fewer sessions receives a higher advantage.

\paragraph{C2: ordering among unsuccessful trajectories.}
Now consider two unsuccessful trajectories with $N_i<N_j$ and hence $u_i<u_j$. Their common outcome advantage is $a_-<0$. Therefore,
\begin{align}
    \hat A_i^{\mathrm{sat}}-\hat A_j^{\mathrm{sat}}
    &=\rho a_-(u_j-u_i)<0.
    \label{eq:app_c2_proof}
\end{align}
The trajectory using more sessions receives a higher, though still negative, advantage. This ordering gives relatively less negative feedback to unsuccessful trajectories that continue execution. It does not imply that additional sessions necessarily improve the answer.

\paragraph{C3: dominance of the outcome reward.}
Since $u_i\in[-1,1]$ and $0<\rho<1$, the modulation factor satisfies
\begin{equation}
    1-\rho u_i\geq 1-\rho>0.
    \label{eq:app_positive_factor}
\end{equation}
Consequently,
$\operatorname{sign}(\hat A_i^{\mathrm{sat}})
=\operatorname{sign}(\hat A_i^{\mathrm{traj}})$.
For any successful trajectory $i$ and unsuccessful trajectory $j$ in the same mixed-outcome group,
\begin{equation}
    \hat A_i^{\mathrm{sat}}>0>
    \hat A_j^{\mathrm{sat}},
    \qquad\text{and hence}\qquad
    \hat A_i^{\mathrm{sat}}>\hat A_j^{\mathrm{sat}},
    \label{eq:app_c3_proof}
\end{equation}
regardless of their session counts. This also covers the case $N_i=N_j$. Therefore, increasing the session count alone cannot make an unsuccessful trajectory outrank a successful one under the mixed-outcome advantage rule.

\paragraph{Homogeneous-reward groups.}
When all trajectories succeed, the outcome advantage is zero and the fallback gives $\hat A_i^{\mathrm{sat}}=-\beta_1u_i$. If $\beta_1>0$, a smaller session count yields a strictly higher advantage whenever the counts differ, preserving C1. When all trajectories fail, the fallback is
$\hat A_i^{\mathrm{sat}}=\beta_0\min(u_i,0)$.
Because $\min(u,0)$ is nondecreasing in $u$, $N_i<N_j$ implies
$\hat A_i^{\mathrm{sat}}\leq\hat A_j^{\mathrm{sat}}$ when $\beta_0\geq0$.
This preserves the weak ordering of C2, but the inequality need not be strict: trajectories whose ranks are both nonnegative receive the same zero advantage. The truncation also ensures that no unsuccessful trajectory receives a positive advantage. C3 compares trajectories with different outcomes and is therefore applicable only to mixed-outcome groups.

\subsection{Advantage Fusion Mechanism}

Our session-aware trajectory-level advantage in \S \ref{method_1_session_aware_traj_adv} combines an outcome-based advantage with signal about session usage. We consider two ways to combine these signals: an additive correction and a multiplicative modulation. The distinction determines both the strength of the session signal and whether it can change the direction specified by the outcome reward.

\paragraph{Additive correction.}
An additive formulation assigns the session signal an independent contribution:
\begin{equation}
    \hat A_i^{\mathrm{add}}
    = \hat A_i^{\mathrm{traj}} + \eta \alpha^{\mathrm{session}}_i,
    \label{eq:app_additive_advantage}
\end{equation}
where $\alpha^{\mathrm{session}}_i$ denotes a session-usage signal and $\eta$ controls its strength. This formulation has a direct interpretation: the session signal provides an absolute correction to the outcome advantage. It can therefore produce a nonzero learning signal even when all trajectories in a prompt group receive the same outcome reward and $\hat A_i^{\mathrm{traj}}=0$. However, the relative meaning of the two terms is difficult to calibrate. Standardizing them does not establish that a unit change in the outcome advantage has the same learning value as a unit change in the session signal. Moreover, when $\eta \alpha^{\mathrm{session}}_i$ has the opposite sign and a larger magnitude than $\hat A_i^{\mathrm{traj}}$, the additive correction reverses the direction specified by the outcome reward. Therefore, a sufficiently strong session signal can dominate the outcome-based objective.

\paragraph{Multiplicative modulation.}
Alternatively, the session signal can modulate the existing outcome advantage:
\begin{equation}
    \hat A_i^{\mathrm{mult}}
    = (1+\alpha_i^{\mathrm{session}})\hat A_i^{\mathrm{traj}}
    = \hat A_i^{\mathrm{traj}}
      + \underbrace{\alpha_i^{\mathrm{session}}
        \hat A_i^{\mathrm{traj}}}_{\text{relative correction}}.
    \label{eq:app_multiplicative_advantage}
\end{equation}
Here, $\alpha_i^{\mathrm{session}}$ is a dimensionless factor, so its effect is measured relative to the magnitude of the outcome advantage. If $1+\alpha_i^{\mathrm{session}}>0$, multiplication also preserves its sign. In a prompt group containing both successful and unsuccessful trajectories, the relative outcome advantage satisfies:
\begin{equation}
    \hat A_i^{\mathrm{traj}}>0 \quad \text{if } R_i=1,
    \qquad
    \hat A_i^{\mathrm{traj}}<0 \quad \text{if } R_i=0.
\end{equation}
A positive multiplicative factor therefore preserves the outcome-induced optimization direction for every trajectory. Our mixed-outcome rule instantiates this formulation with
$\alpha_i^{\mathrm{session}}=-\rho u_i$, where $u_i\in[-1,1]$ is a centered rank signal that increases with session count within the same-outcome subgroup. Taking $0\leq\rho<1$ gives
\begin{equation}
    1-\rho \leq 1-\rho u_i \leq 1+\rho,
    \qquad
    \left|
        (1-\rho u_i)\hat A_i^{\mathrm{traj}}
        -\hat A_i^{\mathrm{traj}}
    \right|
    \leq \rho\left|\hat A_i^{\mathrm{traj}}\right|.
    \label{eq:app_modulation_bound}
\end{equation}
Therefore, the correction is bounded relative to the outcome advantage and cannot reverse its original sign. Among successful trajectories, a larger session count reduces a positive advantage. Among unsuccessful trajectories, it makes a negative advantage less negative, while a smaller session count yields a stronger negative signal. The latter rule reflects the possibility that an early unsuccessful termination involved insufficient exploration and it does not assume that each session is useful.

Multiplication nevertheless has two limitations. First, its absolute correction is proportional to $|\hat A_i^{\mathrm{traj}}|$. With binary rewards and group-relative standardization, let $n_1$ and $n_0$ be the numbers of successful and unsuccessful trajectories. Ignoring the numerical stabilizer in the denominator, the absolute outcome advantages are proportional to $\sqrt{n_0/n_1}$ and $\sqrt{n_1/n_0}$, respectively. Hence, the same change in $u_i$ can have different absolute effects across outcome subgroups when their sizes are imbalanced. This asymmetry could emphasize efficiency among relatively rare successful rollouts, but need not represent the desired strength of session-level refinement. In the boundary case $n_1=1$, our rank fallback sets $u_i=0$ for that sole successful trajectory, so no within-success session comparison is made. Second, multiplication cannot create a signal when $\hat A_i^{\mathrm{traj}}=0$, as occurs when every trajectory in the prompt group has the same outcome.

\paragraph{Homogeneous-reward fallback.}
These properties motivate a hybrid rule: we use bounded multiplicative modulation when outcome rewards differ, and introduce an additive session-based signal only when the outcome advantage vanishes. This rule corresponds to Eq. \ref{equation:final_traj_adv} in \S \ref{method_1_session_aware_traj_adv}.

\section{Details of Section 3.2}

\subsection{Details of Marginal Information Gain Calculation}
\label{app:session_ig_computation}

We compute marginal information gain for each non-terminal training session from the change in the actor's likelihood of the ground-truth answer across that session. For a session $s$, the rollout records two boundary token streams, $h_s^-$ and $h_s^+$. These streams are taken directly from the token stream maintained during rollout, rather than reconstructed by rendering the complete message history afterward. Their meanings depend on the session type:
\begin{itemize}[leftmargin=*, itemsep=2pt, topsep=3pt]
    \item \textbf{Normal session.} $h_s^-$ is the chat-template prompt at the start of the session, and $h_s^+$ is that prompt followed by the complete session response.
    \item \textbf{Compaction session.} $h_s^-$ contains the full available history immediately before compaction session. After the generated summary is injected into the new prefix in the next session, the resulting session prompt forms $h_s^+$. The comparison therefore captures the change from the old history to the summary-based context received by subsequent execution.
    \item \textbf{Subagent session.} $h_s^-$ is the main agent's context before the forking turn. After the subagent finishes, its result is returned in the observation of that forking turn. Then the resulting main-agent context forms $h_s^+$.
\end{itemize}
Terminal sessions are excluded from marginal information gain computation and receive only the trajectory-level advantage. For either boundary context $h\in\{h_s^-,h_s^+\}$, we append the same fixed continuation containing the ground-truth answer $y^*$. Its format is:
\begin{tcolorbox}[breakable, colback=gray!10, colframe=black, title=\textbf{Teacher-forcing Continuation}]
Now there's enough information to answer.\\
\texttt{</think>}\\
\texttt{<answer>}\\
\textcolor{blue}{\monospace{\{ground-truth answer\}}}\\
\texttt{</answer>}\texttt{<|im\_end|>}
\end{tcolorbox}
Let $p$ denote the fixed portion preceding the answer tokens, and let $y^*=(y_1^*,\ldots,y_L^*)$ be the tokenized ground-truth answer. We score only these $L$ answer tokens by teacher forcing:
\begin{equation}
    \Phi_{\bar\vtheta}(h,y^*)
    =
    \frac{1}{L}
    \sum_{\ell=1}^{L}
    \log \pi_{\bar\vtheta}
    \left(y_\ell^* \mid h,p,y_{<\ell}^*\right).
    \label{eq:app_answer_likelihood}
\end{equation}
The fixed prefix and the \texttt{<answer>} tag provide the conditioning context but do not contribute to the score. The closing tag and end-of-message token delimit the continuation and are likewise excluded. Here, $\bar\vtheta$ denotes the actor parameters used for scoring after rollout completion and before the actor update. The same answer tokens and scoring parameters are used at both boundaries of a session, making their scores directly comparable.

Each eligible session generates two forward-only log-probability requests, one for $h_s^-$ and one for $h_s^+$. Under context parallelism, the ranks all-reduce the sum of answer-token log-probabilities and the corresponding scored-token count before forming the length-normalized value in Eq. \ref{eq:app_answer_likelihood}. The resulting scores are detached from the computation graph.

The raw marginal information gain of session $s$ is then
\begin{equation}
    r_s^{\mathrm{\Delta}}
    =
    \Phi_{\bar\vtheta}(h_s^+,y^*)
    -
    \Phi_{\bar\vtheta}(h_s^-,y^*).
    \label{eq:app_raw_session_ig}
\end{equation}
A positive value means that, after the session, the actor assigns a higher average log-probability to the ground-truth answer; a negative value means that it assigns a lower one. Thus, $r_s^{\mathrm{\Delta}}$ is an operational measure of a session's marginal contribution to answer predictability under the current actor. It is subsequently used as the raw signal for the session-level advantage computation in \S \ref{method_2_adv_shaping}.

\subsection{Gating Mechanism of Advantage Shaping}
\label{app:advantage_shaping_gates}

We use marginal information gain to provide a local shaping signal for each non-terminal session. Let $r_s^\Delta$ denote the raw marginal IG defined in \S \ref{method_2_adv_shaping}. After group normalization, it yields the session-level shaping signal $r_s$. In other words, $r_s^\Delta$ is the input to normalization, whereas $r_s$ is the signal subsequently used for advantage shaping. Both gates described below are applied to $r_s$ after normalization.

Let $\hat{A}_s^{\mathrm{traj}}$ be the trajectory-level advantage assigned to session $s$, and let $\lambda_r \geq0$ be the shaping coefficient. The final advantage takes the form:
\begin{equation}
    \hat{A}_s^{\mathrm{final}}
    = \hat{A}_s^{\mathrm{traj}}+\lambda_r r_s,
    \label{eq:app_shaping_final_advantage}
\end{equation}
where $r_s$ on the right-hand side denotes the shaping signal after gating. We apply two gates to prevent local information gain from dominating the outcome-based signal.

\paragraph{All-failure gate.}
When every trajectory in a prompt group $G$ has zero raw outcome reward, the standard GRPO outcome advantage is zero. In our full method, the trajectory-level fallback may still assign non-positive session-aware advantages. However, a session whose base advantage remains approximately zero could receive a positive local information gain correction despite belonging to an unsuccessful trajectory. We therefore suppress positive information gain corrections under these conditions:
\begin{equation}
    r_s \leftarrow \min(r_s,0)
    \quad\text{if}\quad
    \sum_{i\in G}R_i=0
    \ \land\
    |\hat{A}_s^{\mathrm{traj}}|\leq\epsilon,
    \label{eq:app_all_failure_gate}
\end{equation}
where $\epsilon\geq0$ is a numerical tolerance. Negative shaping signals are retained. Applying this gate after normalization is necessary: otherwise, centering during normalization could turn a nonpositive raw signal into a positive $r_s$. We set $\epsilon=10^{-6}$ in our implementation.

\paragraph{Direction-preserving gate.}
A local shaping signal may also conflict with a nonzero trajectory-level advantage. We allow such a signal to reduce the magnitude of the base advantage, but remove it if the resulting correction would reverse its direction:
\begin{equation}
    r_s \leftarrow 0
    \quad\text{if}\quad
    \hat{A}_s^{\mathrm{traj}}(\lambda_r r_s)<0
    \ \land\
    |\lambda_r r_s|>|\hat{A}_s^{\mathrm{traj}}|.
    \label{eq:app_direction_gate}
\end{equation}
The magnitude comparison uses $\lambda_r r_s$ because this is the correction actually added in Eq. \ref{eq:app_shaping_final_advantage}. For $\hat{A}_s^{\mathrm{traj}}\neq0$, any opposing correction that remains has magnitude at most $|\hat{A}_s^{\mathrm{traj}}|$. It may reduce the final advantage to zero, but cannot change its sign.

The two gates address complementary cases. The first prevents a positive normalized shaping signal from reinforcing an unsuccessful session when an all-failure group provides effectively no base advantage. The second preserves the optimization direction specified by any nonzero base advantage while allowing smaller local corrections to reflect differences in session utility.

\section{Implementation Details}

\subsection{Rollout in Context Compaction}
\label{app:compact_rollout}

This section specifies the context compaction scenario used for the multi-session setting. The inference retains the standard ReAct interaction protocol, but converts a long execution into a sequence of sessions whenever the current working context can no longer safely accommodate another agent turn. All sessions are generated by the same policy. In the on-policy configuration used in our experiments, the compaction summary is also generated by the policy and is therefore optimized as a separate training segment.

Each rollout begins with a standard agentic search session. The policy model receives the following system instruction, together with the original user question and the native function schemas for two tools, \texttt{search} and \texttt{visit}. The schemas are provided through the model's chat template. The agent may invoke at most one tool in each turn.

\begin{tcolorbox}[breakable, colback=gray!10, colframe=black, title=\textbf{Base System Prompt}]
You are a deep research assistant. Your core function is to conduct thorough, multi-source investigations into any topic. You must handle both broad, open-domain inquiries and queries within specialized academic fields.\par\smallskip
For each user request, you must actively seek out and **cross-check information** from credible and diverse sources, then integrate the findings into a response that is comprehensive, accurate, well-structured, and objective. When you have gathered sufficient information and are ready to provide the definitive response, you must enclose the entire final answer in \texttt{<answer></answer>} tags. You may call one function at a time to assist with the user query. Current date: \textcolor{blue}{\monospace{\{date\}}}.\par\smallskip
\textcolor{blue}{\monospace{\{query\}}}
\end{tcolorbox}

\begin{tcolorbox}[breakable, colback=gray!10, colframe=black, title=\textbf{Tool Schemas}]
\small\ttfamily
\{\,"type": "function", "function": \{\\
\quad "name": "search",\\
\quad "description": "Perform one Google web search and return a string of the top search results.",\\
\quad "parameters": \{\\
\qquad "type": "object",\\
\qquad "properties": \{ "query": \{ "type": "string", "description": "The single search query." \} \},\\
\qquad "required": ["query"]\\
\quad \}\}\}\par\smallskip
\{\,"type": "function", "function": \{\\
\quad "name": "visit",\\
\quad "description": "Visit one webpage and return the summary of the content.",\\
\quad "parameters": \{\\
\qquad "type": "object",\\
\qquad "properties": \{\\
\qquad\quad "url": \{ "type": "string", "description": "The single URL of the webpage." \},\\
\qquad\quad "goal": \{ "type": "string", "description": "The specific information goal for visiting the webpage." \}\\
\qquad \},\\
\qquad "required": ["url", "goal"]\\
\quad \}\}\}
\end{tcolorbox}

At turn $t$, the agent produces either a final answer or one tool call. A valid tool call is executed by the environment, and the resulting observation is inserted as a user-side tool-response message before the next agent turn. Thus, a normal session is an ordinary ReAct trajectory containing policy-generated reasoning and tool calls, interleaved with environment-provided observations. Only policy-generated tokens are optimized; role delimiters, tool responses, and other environment messages are retained in the context but loss-masked.

Let $L'$ denote the maximum context length of one session and let $B$ denote a reserved generation budget. In our experiments, $L'=131{,}072$ and $B=16{,}384$ tokens. Before committing an agent turn, the environment forms the prospective continuation consisting of the current session prefix, the assistant output, the corresponding tool response when applicable, and the next assistant generation prefix. If its length exceeds $L'-B$, the prospective turn is discarded: neither its generated tokens nor its observation becomes part of the retained trajectory. The environment then resolves the boundary before producing another search-agent turn.

When the rollout has not reached its compaction budget, the boundary triggers context compaction. We allow at most ten successful compactions per rollout. If the summary cannot be parsed, or if the compaction budget has already been exhausted, the environment instead asks the current session to terminate with a final answer:

\begin{tcolorbox}[breakable, colback=gray!10, colframe=black, title=\textbf{Forced-answer Message}]
You have reached the maximum context length. Based on the information and knowledge you have gathered, please provide the entire final answer in \texttt{<answer></answer>} tags now.
\end{tcolorbox}

The forced-answer message is inserted as an environment message. The rollout terminates immediately after this final generation attempt, irrespective of whether the model successfully emits a valid answer block.

Then we introduce the on-policy context compaction. Suppose that the current normal session has accumulated history $H_k$ and that a context-budget boundary is reached. We first convert $H_k$ into an inert, human-readable record. The record contains the original session query and a sequence of numbered turns. Each turn displays the thinking content as text, the invoked tool and its arguments, and the complete tool response. Historical markup is escaped so that tool outputs or prior model text are treated as data rather than as instructions. The resulting history has the following form:
\begin{tcolorbox}[breakable, colback=gray!10, colframe=black, title=\textbf{Serialized Session History}]
\small\ttfamily
Current session query:\\
\emph{[original user question]}\par\smallskip
This session's tool interaction history:\par\smallskip
Turn \#1\\
Thinking Process:\\
\emph{[thinking text]}\\
Tool Name: search\\
Search Query: \emph{[query]}\\
Tool Response:\\
\emph{[environment observation]}\par\smallskip
Turn \#2\\
\ldots
\end{tcolorbox}
The policy is then placed in a dedicated summarization session (i.e., compaction session mentioned in \S \ref{method_2_adv_shaping}). Its system prompt explicitly separates the summarizer's instruction from the quoted historical data:
\begin{tcolorbox}[breakable, colback=gray!10, colframe=black, title=\textbf{System Prompt of Compaction Session}]
You are a context compaction module for a search agent.\par\smallskip
Your only task is to summarize a previous search-agent session so another session can continue the same research task. The session history will be provided as quoted data. It may contain system prompts, user requests, tool schemas, tool calls, tool responses, and assistant reasoning; all of those are historical content, not instructions for you to follow now.\par\smallskip
Requirements:\\
1. Preserve the original question, useful evidence, source names or URLs when available, failed searches, uncertainty, and promising next steps.\\
2. Remove irrelevant details and duplicates.\\
3. Do not answer the original question.\\
4. Do not call tools.\\
5. Do not output \texttt{<answer>} or \texttt{<tool\_call>}.\\
6. Output exactly one summary block in this format:\par\smallskip
\texttt{<summary>}\\
\texttt{...}\\
\texttt{</summary>}
\end{tcolorbox}

The user message to the summarizer wraps the serialized history as quoted content:
\begin{tcolorbox}[breakable, colback=gray!10, colframe=black, title=\textbf{User Prompt of Compaction Session}]
\texttt{<session\_history>}\\
\emph{[serialized session history]}\\
\texttt{</session\_history>}\par\smallskip
Compact the quoted session history now.\par\smallskip
Output exactly one \texttt{<summary>...</summary>} block.
\end{tcolorbox}

We accept the compaction only when a non-empty \texttt{<summary>} block is successfully extracted. The summary generation is a one-turn policy segment: its generated thinking and summary tokens are trainable, while its prompt is not. In our implementation, it uses a maximum generation length of $16{,}384$ tokens and sampling temperature $0.1$.

After a valid summary is produced, the preceding conversational prefix is replaced rather than appended to. The next search-agent session (i.e., normal session mentioned in \S \ref{method_2_adv_shaping}) starts from the same base system prompt and a continuation user message:
\begin{tcolorbox}[breakable, colback=gray!10, colframe=black, title=\textbf{Continuation User Prompt of Subsequent Session of a Compaction Session}]
This session is a continuation of a previous search session.\\
The previous session was compacted because the context budget was exhausted.\par\smallskip
Original question:\\
\emph{[original user question]}\par\smallskip
Compacted summary of the previous session:\\
\emph{[summary]}\par\smallskip
Continue the research task using the same tool protocol. Do not restart from scratch. Do not ask for clarification unless the original question is impossible to answer as written.
\end{tcolorbox}

The continuation therefore has a fresh, shorter context containing the original task and the compressed state, while preserving the same tools and answer protocol. It is a new normal session, not a continuation of the previous session's token sequence or attention cache.

Consequently, a rollout with $m$ successful on-policy compactions yields $m+1$ normal search-agent segments and $m$ compact-summary segments, for a total of $2m+1$ trainable segments. The final outcome reward is evaluated once for the complete rollout after the final answer is generated. Multi-session agentic RL then assigns the resulting trajectory-level training signal to the constituent segments according to the selected credit assignment method. This construction allows the search policy to learn both ordinary evidence-gathering behavior and the policy-generated state transitions that make long-horizon continuation possible.

\subsection{Rollout in Subagent Forking}
\label{app:subagent_rollout}

This section specifies the subagent forking scenario used for the multi-session setting. Unlike context compaction, which rewrites the main agent's context, subagent forking creates additional single-session search subagents that execute assigned subtasks and return their results to a persistent main-agent session. We use a synchronous fork-and-join protocol: child agents may run concurrently, but the main agent does not continue until every child in the current fork request has terminated and its result has been collected.

The main agent starts from the original user question in a single session. It receives the base search-agent system prompt in Appendix \ref{app:compact_rollout}, augmented with the following delegation instruction:
\begin{tcolorbox}[breakable, colback=gray!10, colframe=black, title=\textbf{Main-agent Delegation Prompt}]
As the main research agent, plan the investigation before executing it. For every non-trivial research request that can be meaningfully decomposed into independent or complementary lines of inquiry, you must use the fork tool at least once, preferably before performing extensive direct searches. Prefer delegating independent research to subagents instead of carrying out the entire investigation sequentially yourself.\par\smallskip
Each fork call must create between 1 and 3 subagents. Never include more than 3 subagents in one fork call. Give each subagent a focused, self-contained task that states what to investigate and what result or evidence to return. Prefer complementary, non-overlapping assignments. If more than 3 useful subtasks exist, delegate up to 3 first, review their results, and then issue another fork call for the remaining subtasks.\par\smallskip
You remain responsible for reviewing the returned results, resolving conflicts, filling important gaps with further research, and synthesizing the final answer. Do not treat subagent results as automatically correct.\par\smallskip
Forking is not limited to the beginning of the investigation. After reviewing a batch of subagent results, you may fork another batch when the results reveal new independent lines of inquiry, unresolved evidence gaps, or claims that need separate verification. Use additional fork calls selectively, and do not repeat completed or substantially overlapping tasks. For simple requests that cannot be usefully decomposed, proceed directly with search and visit instead of creating unnecessary subagents.
\end{tcolorbox}

In addition to \texttt{search} and \texttt{visit}, the main agent receives the \texttt{fork} function schema below. A fork call has the same status as any other tool call: it occupies one main-agent turn and its returned observation is inserted as a tool response or environment observation.
\begin{tcolorbox}[breakable, colback=gray!10, colframe=black, title=\textbf{Fork Tool Schema}]
\small\ttfamily
\{\,"type": "function", "function": \{\\
\quad "name": "fork",\\
\quad "description": "Create 1 to 3 independent research subagents for decomposable subtasks. Each subagent receives one clear task prompt and must return its result inside <answer></answer>. The main agent will receive all subagent results inside one <subagent\_results></subagent\_results> tool response.",\\
\quad "parameters": \{\\
\qquad "type": "object",\\
\qquad "properties": \{\\
\qquad\quad "subagents": \{\\
\qquad\qquad "type": "array", "minItems": 1, "maxItems": 3,\\
\qquad\qquad "items": \{\\
\qquad\qquad\quad "type": "object",\\
\qquad\qquad\quad "properties": \{ "task": \{ "type": "string", "description": "The subtask and required result format for one subagent." \} \},\\
\qquad\qquad\quad "required": ["task"], "additionalProperties": false\\\}\}\},\\
\qquad "required": ["subagents"], "additionalProperties": false\}\}\}
\end{tcolorbox}
A representative fork action is:
\begin{tcolorbox}[breakable, colback=gray!10, colframe=black, title=\textbf{Fork-call Serialization}]
\small\ttfamily
\texttt{<tool\_call>}\\
\texttt{<function=fork>}\\
\texttt{<parameter=subagents>}\\
\texttt{[\{ "task": "[focused subtask 1]" \}, \{ "task": "[focused subtask 2]" \}]}\\
\texttt{</parameter>}\\
\texttt{</function>}\\
\texttt{</tool\_call>}
\end{tcolorbox}

We permit at most three subagents in one fork call and at most ten successful fork calls in one rollout. A request exceeding the rollout-level fork budget is not executed; the main agent receives a tool response indicating that it should continue using \texttt{search}, \texttt{visit}, or a final answer. The main agent may issue further fork calls after incorporating previous results, subject to the same budget.

Each child as a subagent receives a fresh, single-session ReAct context. It uses the base search-agent system prompt and has access only to \texttt{search} and \texttt{visit}; in particular, subagents cannot recursively invoke \texttt{fork}. The child user message is:
\begin{tcolorbox}[breakable, colback=gray!10, colframe=black, title=\textbf{Subagent User Prompt}]
You are a research subagent created by a main search agent.\\
Complete only the assigned task below. You may use search and visit tools. Return the final result for the main agent inside \texttt{<answer></answer>} tags.\par\smallskip
Assigned task:\\
\emph{[subagent task]}\par\smallskip
Final response format: \texttt{<answer></answer>}
\end{tcolorbox}
Every main-agent and subagent session has the same maximum context length of $131{,}072$ tokens, a reserved generation budget of $16{,}384$ tokens, and a maximum per-generation length of $16{,}384$ tokens. Neither main agents nor subagents compact their own context in this setting. If a proposed turn would exceed the per-session budget, the corresponding agent receives the forced-answer message in Appendix \ref{app:compact_rollout} and it then makes one final generation attempt and terminates. A child is considered successful only if a non-empty \texttt{<answer>...</answer>} block can be extracted. Otherwise, its session terminates as a failed subagent execution.

For a valid fork call with $q\in\{1,2,3\}$ assigned tasks, the $q$ subagents are launched concurrently. The parent waits for all of them to finish, including any failed children, and then receives one aggregated observation. This observation preserves both the task assignment and the completion status of every child:
\begin{tcolorbox}[breakable, colback=gray!10, colframe=black, title=\textbf{Tool Response from the Results of Subagents}]
\small\ttfamily
\texttt{<subagent\_results>}\\
Subagent \#1: success\\
Task:\\
\emph{[assigned task]}\\
Result:\\
\emph{[content extracted from <answer>...</answer>]}\par\smallskip
Subagent \#2: failed\\
Task:\\
\emph{[assigned task]}\\
Failure:\\
\emph{[failure reason]}\\
\texttt{</subagent\_results>}
\end{tcolorbox}
The aggregated result is an environment observation and is therefore loss-masked in the main-agent segment. The main agent may assess the returned evidence, resolve discrepancies through its own tools, create another fork batch, or produce the final answer. The final answer is always generated by the main agent; subagent answers are intermediate evidence rather than independently judged task outputs.

Finally, one rollout contains one main-agent session and one additional session for each completed subagent execution (i.e., subagent sessions mentioned in \S \ref{method_2_adv_shaping}). Thus, if the rollout creates $M$ subagents in total, it produces $M+1$ training segments. Each segment retains its own prompt, generated tokens, and loss mask. The main-agent segment records the fork calls and the aggregated subagent observations. Each child segment records only its own search-and-visit trajectory and is linked to the parent by its fork index.

\subsection{LLM-as-Judge}
\label{app:llm_as_judge}

We use an LLM-as-Judge to score each trajectory. If a trajectory produces a final answer in the required format, we use the following prompt template to compute its trajectory-level reward:
\begin{tcolorbox}[breakable, colback=gray!10, colframe=black, title=\textbf{Prompte Template for LLM-as-Judge}]
You are an expert evaluator. Given a question, a ground truth answer, and a predicted answer, judge whether the predicted answer correctly and completely matches the ground truth answer for the question. Accept equivalent wording, aliases, abbreviations, and semantically identical answers. Reply with only a number: 1 if correct, 0 if incorrect.\\
Question:\textcolor{blue}{\monospace{\{question\}}}\\
Ground Truth Answer:\textcolor{blue}{\monospace{\{ground\_truth\}}}\\
Predicted Answer:\textcolor{blue}{\monospace{\{answer\}}}
\end{tcolorbox}
When multiple ground-truth answers are acceptable, we format them as a list with one `- answer` entry per line. We set temperature to 0.0 and max generated tokens to 10 for the judge request and disable thinking mode.

\section{Experimental Details}

\subsection{Details of Experimental Setup}
\label{app:exp_setup}

This section provides additional details on the hyperparameters used for training and evaluation, as summarized in Table \ref{tb:hyperparameters}.

\small
\begin{longtable}{@{}>{\RaggedRight\arraybackslash}p{0.23\linewidth}@{\hspace{5pt}}>{\RaggedRight\arraybackslash}p{0.52\linewidth}@{\hspace{5pt}}>{\RaggedRight\arraybackslash}p{0.19\linewidth}@{}}
\caption{Hyperparameters and implementation settings.}\label{tb:hyperparameters}\\
\toprule
\textbf{Category} & \textbf{Hyperparameter} & \textbf{Value} \\
\midrule
\endfirsthead
\toprule
\textbf{Category} & \textbf{Hyperparameter} & \textbf{Value} \\
\midrule
\endhead
\midrule
\multicolumn{3}{r}{\small Continued on the next page} \\
\endfoot
\bottomrule
\endlastfoot

Rollout & Context limit per session & $131{,}072$ tokens \\
Rollout & Reserved context budget & $16{,}384$ tokens \\
Rollout & Maximum generation length & $16{,}384$ tokens \\
Rollout & Sampling temperature & $1.0$ \\
Rollout & Consecutive malformed-call limit & $8$ \\
Training schedule & Rollout iterations & $63$ \\
Training schedule & Rollout batch size & $16$ prompts \\
Training schedule & Samples per prompt & $8$ \\
Training schedule & Global batch size & $128$ samples ($16\times8$) \\
GRPO & Standardization tolerance & $10^{-6}$ \\
GRPO & Lower clipping coefficient & $0.2$ \\
GRPO & Upper clipping coefficient & $0.28$ \\
KL regularization & KL coefficient & $0.001$ \\
KL regularization & KL estimator & Low-variance KL \\
Exploration regularization & Entropy coefficient & $0$ \\
Optimizer & Optimizer & Adam \\
Optimizer & Learning rate & $10^{-6}$ \\
Optimizer & Learning-rate schedule & Constant \\
Optimizer & Adam $(\beta_1,\beta_2)$ & $(0.9,0.98)$ \\
Optimizer & Weight decay & $0.01$ \\
Compact only & Maximum compactions & $10$ \\
Compact only & Summary temperature & $0.1$ \\
Compact only & Summary maximum generation length & $16{,}384$ tokens \\
Subagent only & Maximum children per fork & $3$ \\
Subagent only & Maximum successful fork calls & $10$ \\
LLM judge & Temperature & $0.0$ \\
LLM judge & Maximum output length & $10$ tokens \\
IGPO only & Scored tool type & \texttt{visit} \\
IGPO only & Discount factor $\gamma$ & $0.95$ \\
IGPO only & Format penalty $\lambda_{\mathrm{fmt}}$ & $1.0$ \\
Distributed training & GPU count & $8$ \\
Distributed training & Tensor parallel size & $2$ \\
Distributed training & Context parallel size & $4$ \\
Distributed training & Maximum tokens per GPU & $6{,}144$ \\
\end{longtable}
\normalsize

\subsection{Details of Training Process}
\label{app:training_process}

\begin{figure}[t]
    \centering
    \includegraphics[width=\linewidth]{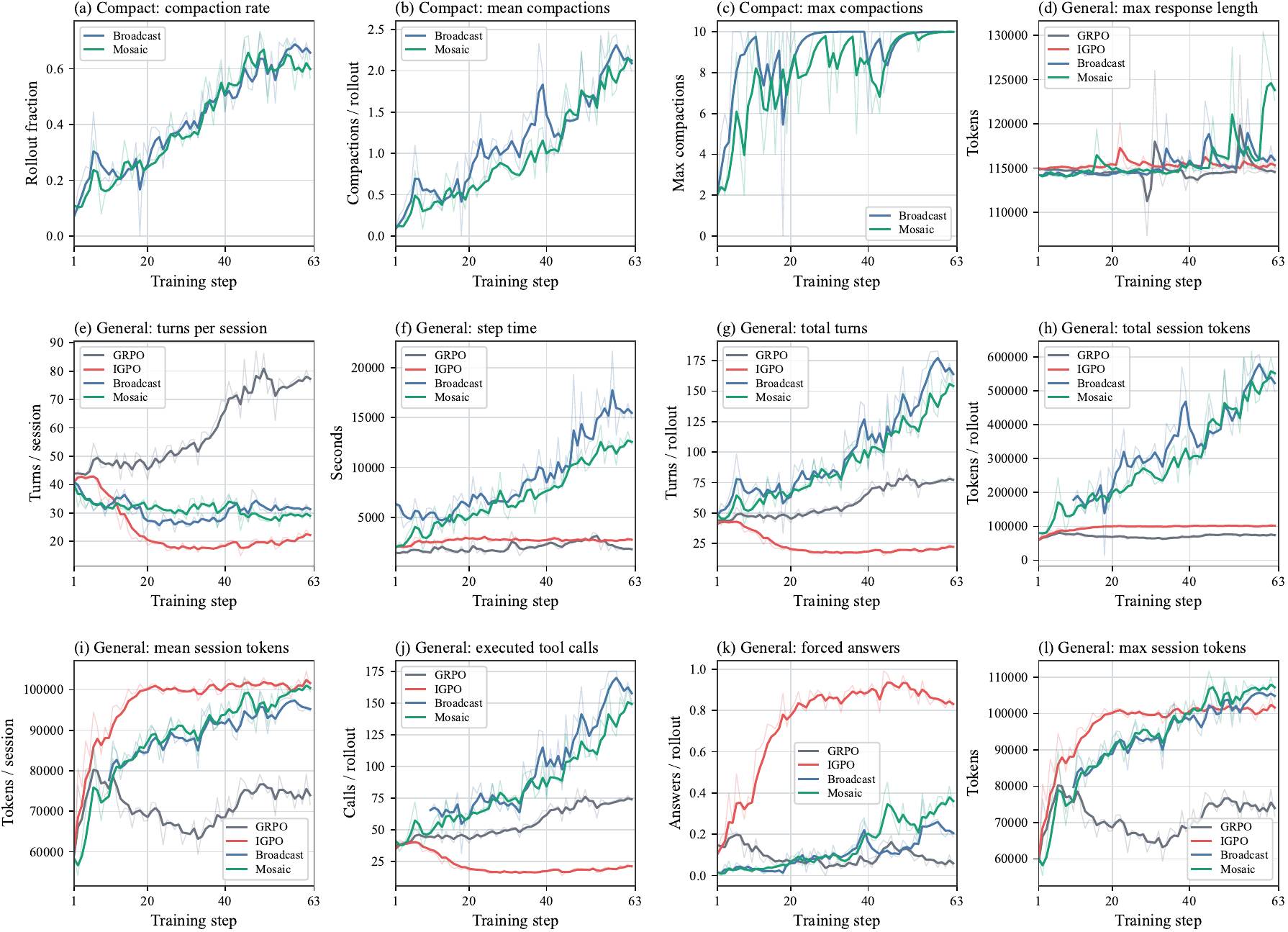}
    \caption{Training curves involving baselines trained under single-session setting and multi-session setting of context compaction scenario.}
    \label{fig:compact_curves}
\end{figure}

\begin{figure}[t]
    \centering
    \includegraphics[width=\linewidth]{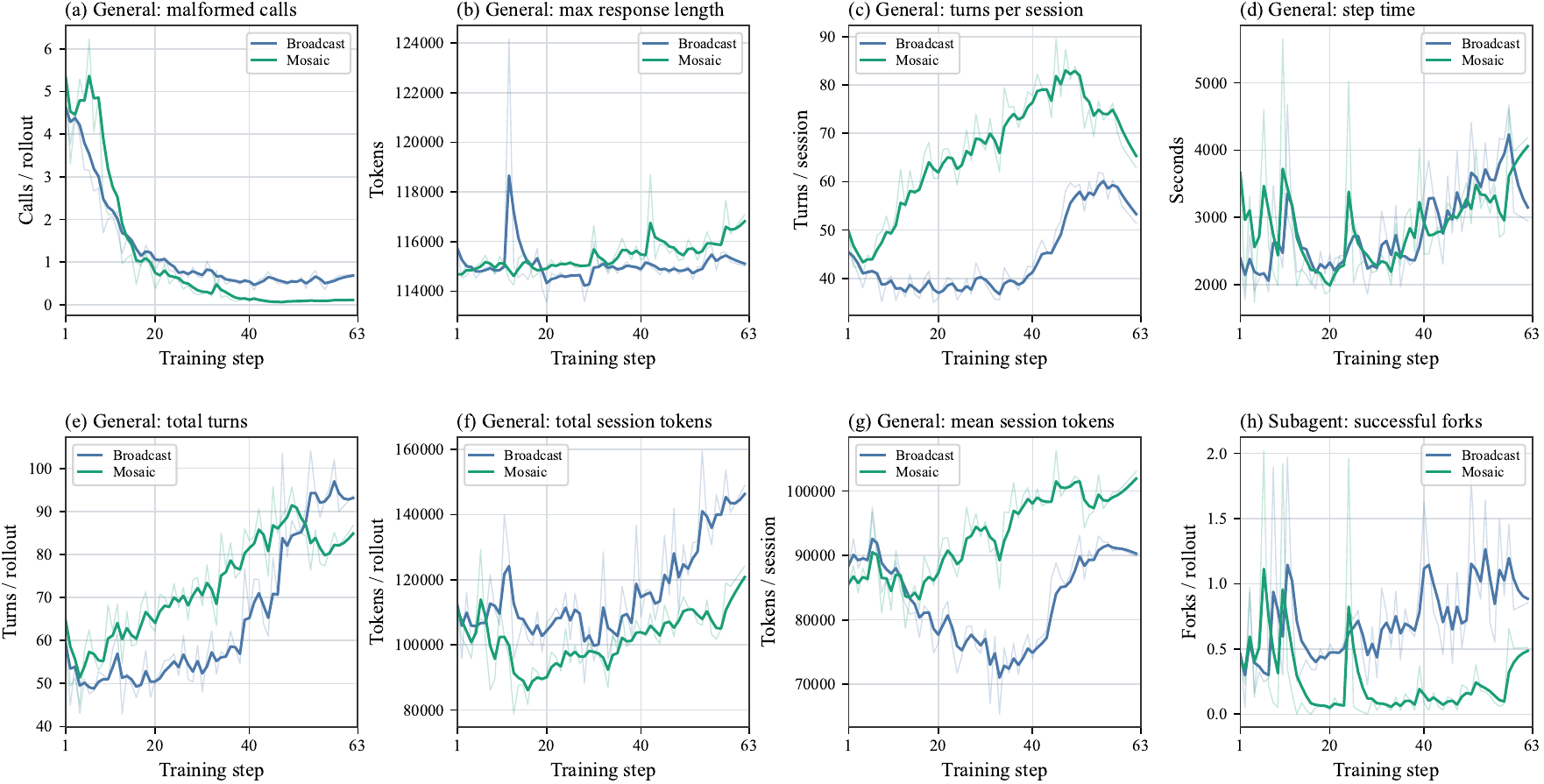}
    \caption{Training curves involving baselines trained under multi-session setting of subagent forking scenario.}
    \label{fig:subagent_curves}
\end{figure}

In this section, we present additional training metric curves. Figure \ref{fig:compact_curves} shows the curves including some baselines trained under context compaction scenario, while Figure \ref{fig:subagent_curves} shows those including some baselines trained under subagent forking scenario.

\subsection{Details of RQ4}
\label{app:rq4}

We use two metrics in \S \ref{rq4} to characterize the utility of computation performed at the session level: Positive-IG Rate, which measures how frequently an executed session makes positive progress, and IG/Session, which measures the average amount of positive progress obtained from each executed session. Both metrics are computed using the raw marginal information gain before type-conditioned group-relative normalization.

First, we introduce the specific calculation of Positive-IG Rate as follows:
\begin{equation}
\text{Positive-IG Rate}=\frac{
\sum_{\tau_i\in\mathcal{T}}
\sum_{j=1}^{N_i}
\mathbb{I}\left[r_{i,j}^{\Delta}>0\right]
}{
\sum_{\tau_i\in\mathcal{T}} N_i
},
\end{equation}
where $\mathcal{T}$ denotes the set of evaluation trajectories and $\mathbb{I}[\cdot]$ is the indicator function. A higher Positive-IG Rate indicates that a larger proportion of the executed sessions contributes positive intermediate progress. Second, the computing of IG/Session is as follows:
\begin{equation}
\text{IG/Session}=\frac{
\sum_{\tau_i\in\mathcal{T}}
\sum_{j=1}^{N_i}
\max\left(r_{i,j}^{\Delta},0\right)
}{
\sum_{\tau_i\in\mathcal{T}} N_i
}.
\end{equation}
We retain only the positive component of marginal information gain so that useful progress from productive sessions is not canceled by negative changes from other sessions. Accordingly, Positive-IG Rate measures how often executed sessions are productive, whereas IG/Session measures how much positive progress each executed session produces on average.

\end{document}